\documentclass[sigconf]{acmart}
\usepackage{amsmath,amsfonts}
\usepackage{algorithmic}
\usepackage{algorithm}
\usepackage{array}
\usepackage[caption=false,font=normalsize,labelfont=sf,textfont=sf]{subfig}
\usepackage{textcomp}
\usepackage{stfloats}
\usepackage{caption}
\usepackage{url}
\usepackage{verbatim}
\usepackage{multirow}
\usepackage{graphicx}
\usepackage{orcidlink}
\usepackage{cleveref}
\usepackage{bm}
\usepackage{balance}

\AtBeginDocument{%
  }

\setcopyright{acmlicensed}
\copyrightyear{2025}
\acmYear{2025}
\setcopyright{cc}
\setcctype{by}
\acmConference[MM '25]{Proceedings of the 33rd ACM International Conference
on Multimedia}{October 27--31, 2025}{Dublin, Ireland}
\acmBooktitle{Proceedings of the 33rd ACM International Conference on
Multimedia (MM '25), October 27--31, 2025, Dublin,
Ireland}\acmDOI{10.1145/3746027.3758240}
\acmISBN{979-8-4007-2035-2/2025/10}

\begin{document}

\title{MathScape: Benchmarking Multimodal Large Language Models in Real-World Mathematical Contexts}


\author{Hao Liang$^\dagger$}
\affiliation{%
  \institution{Peking University}
  \city{Beijing}
  \country{China}}
\email{hao.liang@stu.pku.edu.cn}

\author{Linzhuang Sun$^\dagger$}
\affiliation{%
  \institution{University of Chinese Academy of Sciences}
  \city{Beijing}
  \country{China}
}
\email{sunlinzhuang21@mails.ucas.ac.cn}

\author{zhouminxuan$^\dagger$}
\affiliation{%
  \institution{Nankai University}
  \city{Tianjin}
  \country{China}
}
\email{2120220674@mail.nankai.edu.cn}

\author{Zirong Chen}
\affiliation{%
  \institution{Beijing Institute of Technology}
  \city{Beijing}
  \country{China}
}
\email{1120223580@bit.edu.cn}

\author{Meiyi Qiang}
\affiliation{%
  \institution{Peking University}
  \city{Beijing}
  \country{China}
}
\email{qiangmeiyi@gmail.com}

\author{Mingan Lin}
\affiliation{%
  \institution{Baichuan Inc.}
  \city{Beijing}
  \country{China}
}
\email{mingan5547@gmail.com}

\author{Tianpeng Li}
\affiliation{%
  \institution{Baichuan Inc.}
  \city{Beijing}
  \country{China}
}
\email{litianpeng@baichuan-inc.com}

\author{Fan Yang}
\affiliation{%
  \institution{Baichuan Inc.}
  \city{Beijing}
  \country{China}
}
\email{yangfan@baichuan-inc.com}

\author{Zenan Zhou$^*$}
\affiliation{%
  \institution{Baichuan Inc.}
  \city{Beijing}
  \country{China}
}
\email{zenanchow@gmail.com}

\author{Wentao Zhang$^*$}
\affiliation{%
  \institution{Peking University}
  \city{Beijing}
  \country{China}
}
\email{wentao.zhang@pku.edu.cn}
\renewcommand{\shortauthors}{Hao Liang et al.}

\begin{abstract}
With the rapid progress of Multimodal LLMs, evaluating their mathematical reasoning capabilities has become an increasingly important research direction. In particular, visual-textual mathematical reasoning serves as a key indicator of an MLLM’s ability to comprehend and solve complex, multi-step quantitative problems. While existing benchmarks such as MathVista and MathVerse have advanced the evaluation of multimodal math proficiency, they primarily rely on digitally rendered content and fall short in capturing the complexity of real-world scenarios. To bridge this gap, we introduce MathScape, a novel benchmark focused on assessing MLLMs’ reasoning ability in realistic mathematical contexts. MathScape comprises 1,369 high-quality math problems paired with human-captured real-world images, closely reflecting the challenges encountered in practical educational settings. We conduct a thorough multi-dimensional evaluation across nine leading closed-source MLLMs, three open-source MLLMs with over 20 billion parameters, and seven smaller-scale MLLMs. Our results show that even SOTA models struggle with real-world math tasks, lagging behind human performance—highlighting critical limitations in current model capabilities. Moreover, we find that strong performance on synthetic or digitally rendered images does not guarantee similar effectiveness on real-world tasks. This underscores the necessity of MathScape in the next stage of multimodal mathematical reasoning. 
\end{abstract}

\ccsdesc[300]{Computing methodologies~Image and video acquisition}
\keywords{Real-World Benchmark, Multimodal Mathematical Benchmark}


\maketitle
\begingroup
\renewcommand\thefootnote{}\footnote{\noindent
$\dagger$ The first three authors have equal contributions. \\
$*$ Corresponding Authors.
}
\addtocounter{footnote}{-1}
\endgroup

\begin{figure}[h!]
  \centering
  \includegraphics[width=0.48\textwidth]{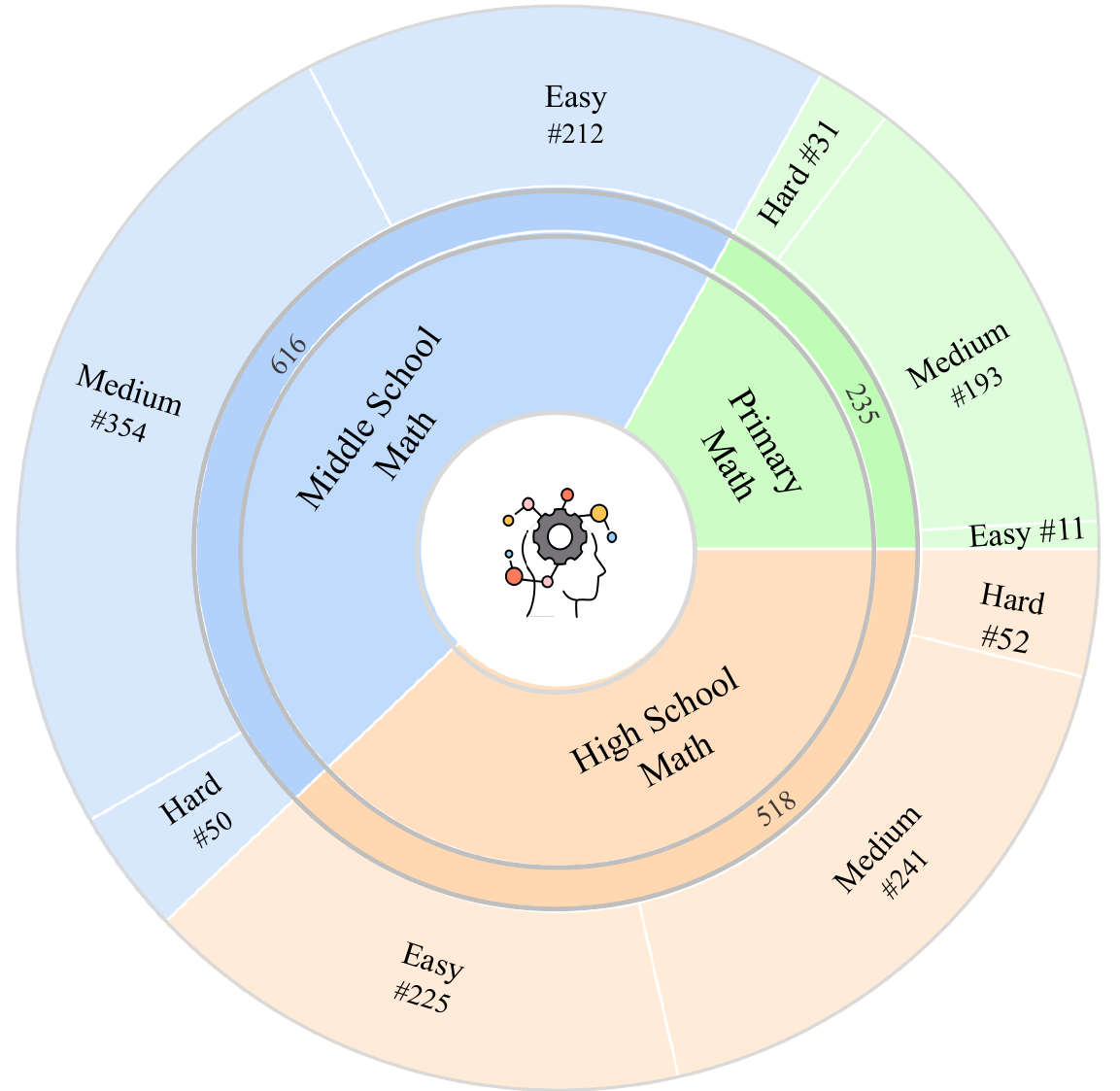}
  \caption{MathScape offers a comprehensive collection of math problems from primary school to high school. The problems range in difficulty from easy to difficult, catering to various levels of evaluation.}
  \label{fig: Face_1}
\end{figure}
\section{Introduction}
Large language models (LLMs) have demonstrated exceptional performance across diverse tasks spanning myriad domains~\cite{chatgpt, llama}. Based on LLMs, MLLMs~\cite{zhao2023survey, wu2023multimodal, bai2024survey,an2025unictokens,an2024mc,luo2024llm} also show strong understanding ability among different modalities~\cite{llava, bai2023qwenvl}.
Among MLLMs, Vision-Language Models (VLMs) have demonstrated state-of-the-art performance in traditional multimodal tasks, including image classification~\cite{chen2024internvl}, image understanding~\cite{blip, li2023blip}, and image captioning~\cite{bai2023qwenvl}. Furthermore, their strong language understanding capabilities enable high performance in text-rich tasks, such as visual question answering~\cite{llava,llava1.5} and image-text retrieval~\cite{chen2024internvl}. Recently, MLLMs have also made significant strides in solving mathematical problems. Therefore, a well-designed benchmark is essential to systematically evaluate their mathematical reasoning abilities. Several benchmarks, such as MATH-V~\cite{wang2024measuring}, MathVerse~\cite{zhang2024mathverse}, and MathVista~\cite{lu2023mathvista}, have been developed to assess the mathematical capabilities of MLLMs. While these works represent notable progress, existing benchmarks primarily rely on synthetic or digitally rendered mathematical problems. In contrast, real-world users often pose math questions based on photos of printed documents or screens, introducing additional challenges such as image quality variations and contextual ambiguity. To bridge this gap, it is essential to develop a comprehensive benchmark that better reflects real-world scenarios.

\begin{figure*}[t]
\centering
\includegraphics[width=\textwidth]{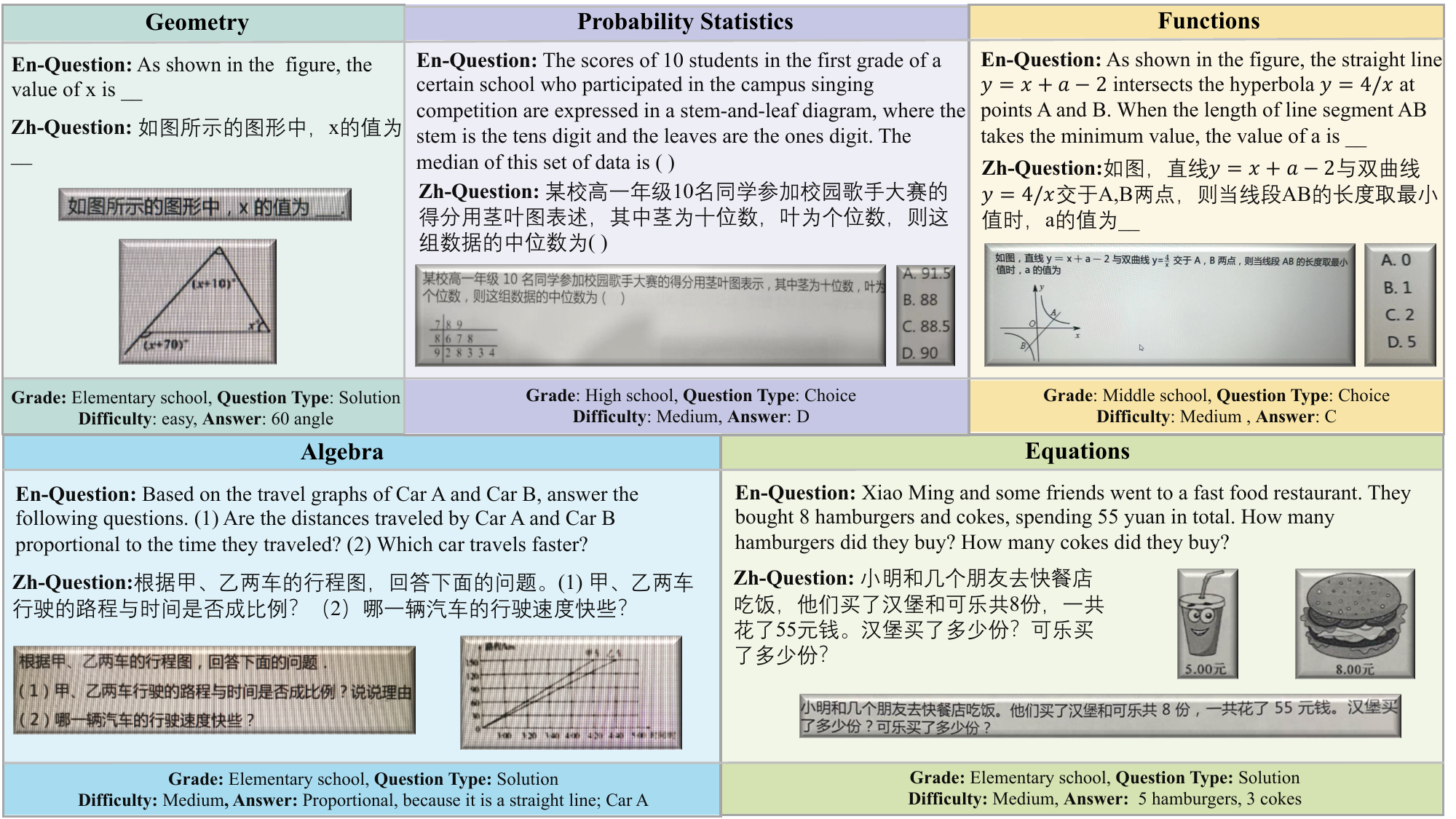} 
\caption{MathScape Data Illustration. We select real-world mathematics examples from various mathematical domains, including geometry, probability and statistics, functions, algebra, and equations. 
}
\label{fig:Main_pipeline}
\end{figure*}

In this paper, we introduce MathScape, a \textbf{real-world-focused} benchmark designed to assess the mathematical problem-solving abilities of MLLMs in practical settings. We implement a three-step pipeline for constructing a real-world math image dataset. The construction process begins with converting mathematical documents into images. Next, we capture photos and screenshots to simulate real-world data collection. Finally, we conduct rigorous review and knowledge classification to ensure high dataset quality. Our MathScape data examples are shown in Figure \ref{fig:Main_pipeline}. For the benchmark assessment, we propose a two-step pipeline tailored to longer mathematical problems. First, we use LLMs to extract answers for each subproblem. Then, we employ LLMs as evaluators to assess the correctness of each solution. To ensure benchmark quality, we incorporate human evaluations at every stage. Through this data construction and evaluation pipeline, we develop MathScape, a multimodal dataset that integrates real-world math problem photos with verified solutions, establishing a more realistic evaluation setting for MLLMs.

The core contributions are summarized as follows:
\begin{itemize}
    \item \textbf{New Benchmark:} We introduce MathScape, a novel multimodal math benchmark designed for real-world mathematical problem-solving. This benchmark comprises 1,369 newly curated data points covering diverse multimodal mathematical scenarios.

    \item \textbf{Detailed Classification Criteria:} We categorize all 1,369 problems according to question type, knowledge domain, and educational level. This fine-grained classification ensures that our benchmark can comprehensively evaluate models across diverse aspects of mathematical understanding. Detailed statistics are shown in Tables~\ref{tab:merged_1_2} and~\ref{Table 3}.

    \item \textbf{Comprehensive Evaluation:} We perform an extensive evaluation on the MathScape benchmark, covering a broad spectrum of models: nine widely adopted closed-source MLLMs, three open-source MLLMs with over 20 billion parameters, and seven smaller-scale models, including two specifically tailored for mathematics. For comparison, we also include random selection, frequency-based prediction, and human performance baselines.
    
    \item \textbf{Real-World Math Problem Evaluation:} By comparing the performance of GPT-4o (PDF) and LLaVA-OneVision-72B (PDF) on digitally rendered versus real-world images in Table \ref{tab:merged_1_2}, and \ref{Table 3}, we demonstrate that real-world mathematical scenarios introduce additional reasoning challenges. This highlights the unique contribution of our benchmark in utilizing real-world images for evaluation.

\end{itemize}
\section{Related Work}

\subsection{Benchmarks for Mathematical Evaluation}
Recent research has made substantial progress in developing benchmarks to evaluate mathematical reasoning abilities. In this section, we review multimodal math benchmarks.

\textbf{Multimodal Benchmarks}
With the rapid advancement of MLLMs, several high-quality benchmarks have been introduced to assess mathematical problem-solving in visual contexts. MathVista~\cite{lu2023mathvista} focuses on visual math QA tasks, evaluating model performance across various mathematical domains, including arithmetic and algebra, within visually rich scenarios. MATH-V~\cite{wang2024measuring} is another benchmark designed to assess multimodal mathematical understanding, with questions primarily sourced from math competitions. MathVerse~\cite{zhang2024mathverse} evaluates MLLMs' comprehension of visual diagrams by employing Chain-of-Thought (CoT) reasoning on 2,612 multimodal math problems. CMMU~\cite{he2024cmmu} is a large-scale Chinese benchmark for multi-disciplinary multimodal understanding, incorporating questions from college exams and textbooks.

\section{Methodology}

\subsection{Construction of MathScape}\label{sec: construction}

\quad\textbf{Data Preparation}
The data preparation involves three main steps. First, we collected a total of 1,369 mathematics problems, each potentially containing \textbf{multiple sub-questions}, from primary, middle, and high school exams and homework. These questions are sourced from primary and secondary school education question banks in China, with difficulty levels annotated as easy, medium, and hard to support fine-grained evaluation. Next, the question documents were converted into PDF format using Pandoc. Finally, the PDFs were rendered into images for subsequent processing.

\textbf{Visual Data Simulation}
After collecting the data, the images are then transformed to closely align with real-world scenarios by capturing photos of printed images and screen displays.

\textbf{Manual Validation}  
After constructing the dataset, we performed a rigorous validation process, costing approximately \$8,000 in labor. We hired five mathematics graduate students from top-tier universities to verify and review each question and answer until consensus was reached. Additionally, both textual and graphical inputs were carefully checked for clarity and accuracy.


\textbf{Knowledge-Based Classification}
Once data quality is verified, the dataset is classified by knowledge points according to criteria grounded in the structure of mathematical knowledge and refined through expert discussion. The classification is conducted by three annotators, followed by verification and organization by two additional reviewers to ensure accuracy and consistency.


\subsection{Multidimensional Evaluation}\label{sec: Evaluation}
To comprehensively assess the performance of MLLMs, we designed a multidimensional evaluation framework to classify and analyze their mathematical capabilities across various categories. The classification criteria are as follows:

\textbf{Question Types} 
We categorized the test questions into multiple formats, including multiple-choice, fill-in-the-blank (solution), and proof-based questions, to examine the model's performance across different question structures.

\textbf{Knowledge Points} 
We further classified the questions based on mathematical knowledge domains, such as algebra, geometry, probability, and statistics, to evaluate the model's proficiency in different areas of mathematics.

\textbf{Educational Stages} 
Additionally, we grouped the questions according to educational levels—primary school, middle school, and high school—to assess the model's adaptability and accuracy across varying levels of complexity.

Each question is independently labeled by three annotators. In cases of disagreement, two adjudicators review the annotation and resolve conflicts through discussion. This multi-stage annotation process ensures the consistency and reliability of the classification.

\subsection{Evaluation Method}\label{sec: Evaluation2}
We employ a two-step process to score long answers effectively.

\textbf{Answer Segmentation} As illustrated in Figure 1 in the appendix, we prompt the LLMs to decompose lengthy answers into multiple sub-answers, each focusing on a specific aspect of the problem. This segmentation ensures that complex responses are broken down into manageable components, facilitating the evaluation of correctness and relevance. By isolating sub-problems within the overall solution, we achieve a more granular analysis of the model's performance.

\textbf{Sub-Answer Scoring} After segmenting the long answer, we utilize the prompt depicted in Figure 2 in the Appendix to automatically score each sub-answer individually. This evaluation strategy allows us to assess the accuracy and completeness of each individual component, ensuring that the final aggregated score offers a comprehensive reflection of the model’s ability to tackle different facets of the original problem. By evaluating sub-answers separately, we identify specific areas where the model excels or struggles, providing deeper insights into its strengths and weaknesses.

To validate the effectiveness of our evaluation method, three annotators manually reviewed the assessments of the MathScape dataset. The automated results were found to align with human judgment in over 97\% of cases, demonstrating the reliability and effectiveness of our evaluation approach.

\subsection{Statistical Characteristics of MathScape}\label{sec: Statistics}

\begin{figure}
\centering
\includegraphics[width=0.48\textwidth]{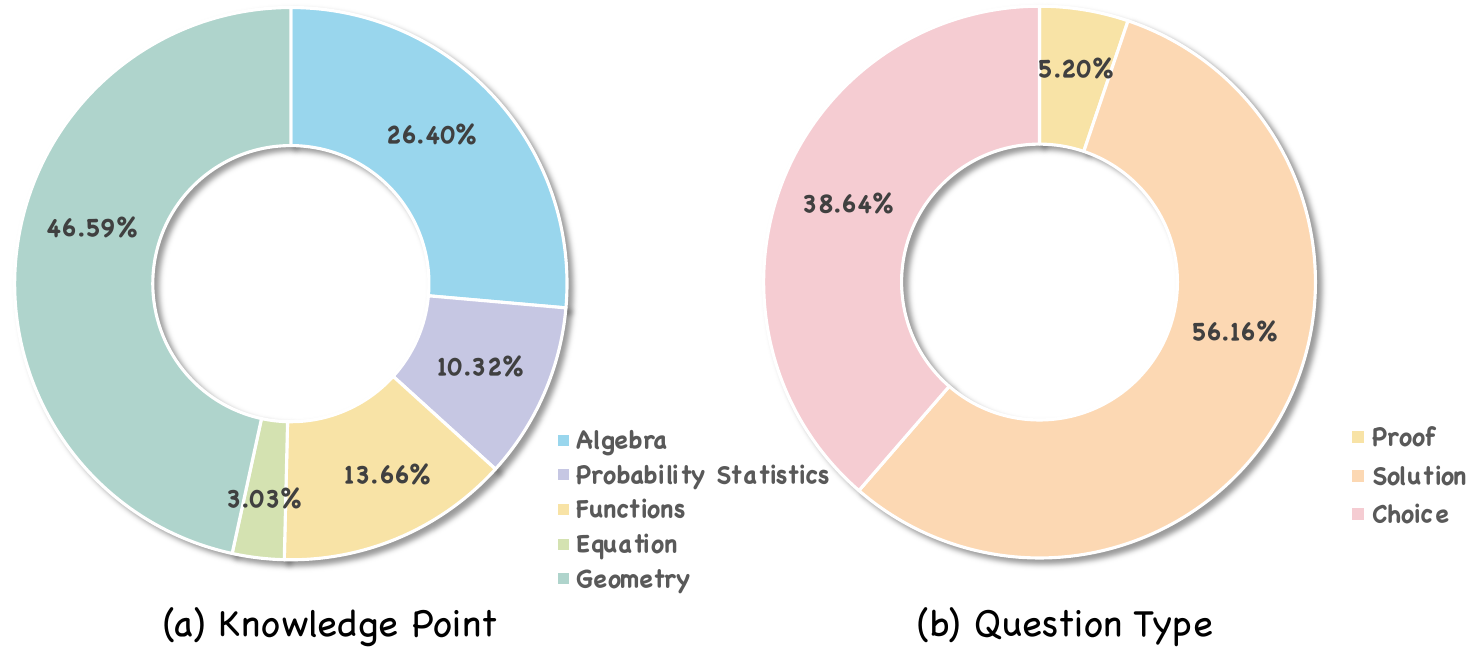} 
\caption{The distribution of MathScape. In (a), we show the proportion based on knowledge areas, while in (b), we present the proportion based on question types.}
\label{fig:proportion}
\vspace{-4mm}
\end{figure}

In this section, we present a summary of the statistical properties of the MathScape dataset. The dataset comprises 1,369 problems, each potentially containing multiple sub-questions, accompanied by corresponding labels, attribute information, detailed solution processes, and standard reference answers. The overall statistics are illustrated in Figures~\ref{fig: Face_1} and~\ref{fig:proportion}. As shown in Figure~\ref{fig: Face_1}, the dataset is categorized by educational stage and difficulty level. We observe a higher concentration of problems originating from middle and high school levels, while primary school problems are relatively underrepresented. In terms of difficulty, the majority of questions fall within the medium level, reflecting their prevalence in real-world educational assessments.

Figure~\ref{fig:proportion} (a) highlights the multimodal nature of the dataset, which integrates both textual and visual elements. A substantial portion of the problems involve geometry, a domain that often necessitates the use of diagrams for effective reasoning. In contrast, topics such as equations and inequalities appear less frequently, consistent with the typical distribution in multimodal problem settings. Figure~\ref{fig:proportion} (b) shows the distribution of question types. The dataset predominantly consists of solution-based and multiple-choice questions, with proof-based questions comprising a smaller fraction. This composition reflects the goal of assessing models on a diverse range of question formats while prioritizing practical problem-solving scenarios that mirror real-world educational tasks.

\section{Experiments and Analysis}\label{sec:Experimental Results}
In this section, we present a comprehensive set of experiments designed to address the following research questions:
\textbf{Q1}: Does our benchmark present substantial challenges for current SOTA MLLMs?
\textbf{Q2}: To what extent do real-world images affect the performance of SOTA MLLMs?
\textbf{Q3}: How consistent and stable are model predictions on our benchmark?
\subsection{Experimental Setups}
\quad\textbf{Models}
In our evaluation of multimodal LLMs, we focused on both open-source and closed-source models that rank among the top performers on major multimodal LLM leaderboards. We included 9 closed-source models, including GPT-4V (gpt-4-1106-preview), GPT-4-Turbo (gpt-4-turbo-2024-04-09)~\cite{openai2023gpt}, GPT-4o (gpt-4o-2024-08-06)~\cite{openai2023gpt}, GeminiPro~\cite{reid2024gemini}, Claude-3-Opus, Baichuan-VL ~\cite{yang2023baichuan}, Qwen-Max~\cite{bai2023qwen}, Qwen-Plus ~\cite{bai2023qwen}, GLM4V~\cite{glm2024chatglm}.
For open-source models, we evaluate 3 large models (>20B parameters) such as Yi-VL-34B~\cite{young2024yi}, Qwen2-VL-Instruct-72B~\cite{bai2023qwenvl}, LLaVA-OneVision-72B (llava-onevision-qwen2-72b-ov-hf)~\cite{li2024llava}. Additionally, we evaluate 7 commonly used small open-source models, including Deepseek-VL2-4.5B~\cite{lu2024deepseek}, LLaVA-1.6-7B\cite{liu2024visual} Qwen2-VL-Instruct-7B~\cite{bai2023qwenvl}, LLaVA-OneVision-7B (llava-onevision-qwen2-7b-ov-hf)~\cite{li2024llava} and Llama-3.2-11B-Vision~\cite{llama} and two math-specific MLLMs Math-LLaVA~\cite{shi2024math} and G-LLaVA-7B~\cite{gao2023g}.
\begin{table*}[t]
\centering
\caption{\textbf{Accuracy comparison of models across question types and knowledge points}}
\label{tab:merged_1_2}
\resizebox{\textwidth}{!}{%
\begin{tabular}{lcccc|ccccc}
\toprule
\textbf{Model} & \textbf{Avg} & \textbf{Choice} & \textbf{Solution} & \textbf{Proof} 
& \textbf{Algebraic} & \textbf{Geometric} & \textbf{Equations} & \textbf{Functions} & \textbf{Probability/Stats} \\
\midrule
Random & 9.78 & 25.39 & 0.0 & 0.0 
& 8.60 & 8.77 & 16.67 & 9.29 & 16.24 \\
Frequency & 9.78 & 26.19 & 0.0 & 0.0 
& 9.17 & 9.15 & 17.50 & 8.60 & 13.52 \\
Human & \textbf{76.96} & \textbf{91.94} & \textbf{71.48} & \textbf{28.99} 
& \textbf{88.61} & \textbf{85.16} & \textbf{80.00} & \textbf{91.40} & \textbf{72.22} \\
\midrule
\multicolumn{10}{c}{\textbf{Closed-source Models}} \\
\midrule
GPT-4V & 34.96 & 35.75 & 31.72 & 28.33 
& 39.05 & 27.90 & 29.73 & 34.14 & 41.31 \\
GPT-4o & 42.47 & 45.72 & 38.16 & \textbf{66.79} 
& 50.58 & \textbf{39.74} & \textbf{38.75} & 35.08 & 44.78 \\
GPT-4o (PDF) & \textbf{43.89} & \textbf{45.81} & \textbf{42.05} & 50.48 
& \textbf{53.82} & 38.60 & 32.08 & \textbf{40.28} & \textbf{50.50} \\
GPT-4-turbo & 33.92 & 29.85 & 31.58 & 56.62 
& 36.28 & 29.54 & 32.50 & 28.43 & 37.99 \\
Claude-3-Opus & 28.79 & 29.30 & 20.85 & 50.00 
& 31.78 & 22.67 & 20.83 & 20.58 & 36.22 \\
Gemini-Pro & 21.37 & 12.62 & 16.16 & 37.50 
& 21.13 & 15.50 & 15.35 & 9.57 & 13.33 \\
Baichuan-VL & 30.00 & 26.38 & 25.83 & 45.97 
& 30.54 & 25.98 & 25.83 & 26.69 & 23.67 \\
Qwen-VL-Max & 27.83 & 23.97 & 22.17 & 34.85 
& 28.71 & 21.86 & 28.33 & 20.86 & 19.09 \\
Qwen-VL-Plus & 15.60 & 19.46 & 12.48 & 35.19 
& 16.70 & 17.07 & 18.67 & 16.67 & 11.46 \\
GLM4V & 12.26 & 11.54 & 7.31 & 26.28 
& 8.94 & 12.57 & 5.13 & 7.32 & 10.55 \\
\midrule
\multicolumn{10}{c}{\textbf{Open-source Models (> 20B)}} \\
\midrule
Qwen2-VL-72B & \textbf{38.67} & \textbf{46.25} & \textbf{34.07} & 33.70 
& \textbf{44.62} & \textbf{34.77} & 33.77 & \textbf{37.25} & \textbf{43.93} \\
LLaVA-Onevision-72B & 8.31 & 7.96 & 8.01 & 14.37 
& 7.87 & 8.01 & 5.98 & 9.77 & 9.49 \\
LLaVA-Onevision-72B(PDF) & 30.56 & 35.08 & 26.88 & \textbf{37.44} 
& 33.19 & 30.08 & \textbf{34.26} & 25.64 & 31.62 \\
Yi-VL-34B & 18.36 & 19.01 & 9.98 & 33.33 
& 16.78 & 15.84 & 7.02 & 9.79 & 11.44 \\
\midrule
\multicolumn{10}{c}{\textbf{Open-source Models (< 20B)}} \\
\midrule
Qwen2-VL-7B & \textbf{26.74} & \textbf{27.76} & \textbf{23.36} & \textbf{57.13} 
& \textbf{34.04} & \textbf{23.89} & \textbf{21.25} & \textbf{25.00} & \textbf{24.92} \\
Llama-3.2-11B-Vision & 20.83 & 25.30 & 15.07 & 52.29 
& 22.54 & 21.44 & 20.42 & 18.82 & 16.57 \\
LLaVA-Onevision-7B & 16.68 & 22.78 & 10.62 & 41.29 
& 12.44 & 20.56 & 14.53 & 15.15 & 12.67 \\
Math-LLaVA-13B & 15.77 & 17.97 & 10.90 & 55.05 
& 11.40 & 19.71 & 15.28 & 12.57 & 13.89 \\
DeepSeek-VL2-4.5B & 15.66 & 12.75 & 10.60 & 37.69 
& 12.71 & 14.87 & 6.19 & 10.60 & 9.61 \\
LLaVA-1.6-7B & 12.35 & 11.31 & 6.24 & 13.43 
& 9.76 & 8.58 & 15.79 & 3.57 & 10.77 \\
G-LLaVA-7B & 13.58 & 27.31 & 5.34 & 6.86 
& 11.42 & 13.58 & 4.58 & 18.32 & 15.32 \\
\bottomrule
\end{tabular}
}
\vspace{-2mm}
\end{table*}

\textbf{Evaluation} We evaluate all the aforementioned models on the MathScape benchmark. Additionally, we select two models—GPT-4o (gpt-4o-2024-08-06)~\cite{openai2023gpt} and LLaVA-OneVision-72B (llava-onevision-qwen2-72b-ov-hf)~\cite{li2024llava}—and assess their performance using a pure PDF file, without images captured by humans (which were intentionally included to increase task difficulty). These results highlight that strong performance on PDF-based inputs does not necessarily translate to human-captured images, underscoring the significance of our MathScape benchmark. Additional results on pure PDF inputs can be found in the Appendix. For the generated answers, we employ GPT-4V to segment and evaluate the responses produced by the selected models.

\textbf{Settings}
We conduct all model inferences in a zero-shot setting, using the same configuration for each official model. 
The settings include a max token limit of 2048, top-k of 5, a temperature of 0.3, and a repetition penalty of 1.05. All experiments are conducted on NVIDIA H100 GPUs.



\subsection{Performance of Various Models}
To address \textbf{Q1}, we evaluate widely adopted MLLMs on the MathScape benchmark and conduct a detailed analysis across question types, knowledge domains, and educational stages.

As shown in Tables \ref{tab:merged_1_2} and \ref{Table 3}, GPT-4o, along with GPT-4o (PDF), consistently achieves the highest accuracy across various question types, knowledge domains, and educational stages. Among open-source models, Qwen2-VL excels across all benchmarks, owing to its innovative architecture and comprehensive training data.

Moreover, the results in Table \ref{tab:merged_1_2} indicate that models generally perform better on proof questions than on multiple-choice and solution-based questions. This suggests that the structured format and clearer information in proof questions make them easier for models to process, while solution-based questions, which often require extensive calculations and complex multi-step reasoning, pose a greater challenge for MLLMs. From Table \ref{tab:merged_1_2}, we also observe that MLLMs perform well on algebraic questions, likely due to their strong capability in solving algebra problems. However, as shown in Table \ref{Table 3}, MLLMs' performance deteriorates as problem difficulty and educational stage increase, indicating that more advanced concepts and problem types are more challenging for these models.

Additionally, mathematical models such as Math-LLaVA and G-LLaVA-7B lag behind more generalized models like Qwen2-VL, highlighting the ability of MLLMs to acquire both general and mathematical knowledge. LLaVA-Onevision, trained exclusively on open-source data, performed well but still fell short of Qwen2-VL, demonstrating the need for more advanced open-source multimodal math datasets.

Overall, these findings highlight the growing strength of general-purpose models in addressing both general and mathematical domains, with clear evidence that the quality and scope of training data, along with model architecture, significantly influence performance across different question types and knowledge domains.

\begin{table*}
\caption{\textbf{Model performance on different educational stages} (\textbf{E: Easy, M: Medium, H: Hard, Avg: Average Score})}
\label{Table 3}
\centering
\resizebox{0.93\textwidth}{!}{%
\begin{tabular}{l c ccc c ccc c ccc}
\toprule
\textbf{Model} & \multicolumn{4}{c}{\textbf{Primary School}} & \multicolumn{4}{c}{\textbf{Middle School}} &\multicolumn{4}{c}{\textbf{High School}}  \\ 
\midrule
 & avg & E & M & H & avg & E & M & H & avg & E & M & H  \\ 
\midrule
Frequency & 5.58 & 9.09 & 4.71 & 9.68 & 11.13 & 15.57 & 9.19 & 6.00 & 10.08 & 13.78 & 7.08 & 7.84 \\
Random & 4.48 & 0.00 & 4.92 & 3.33 & 9.95 & 12.02 & 9.38 & 4.55 & 12.02 & 14.55 & 10.34 & 8.51 \\
Human & \textbf{89.36} & \textbf{100.0} & \textbf{90.16} & \textbf{80.65} & \textbf{84.58} & \textbf{95.75} & \textbf{82.20} & \textbf{54.00} & \textbf{62.21} & \textbf{78.57} & \textbf{51.67} & \textbf{40.38}\\
\midrule
\multicolumn{13}{c}{\textbf{Closed-source Models}} \\
\midrule
GPT-4V & 36.04 & 57.58 & 38.64 & 10.71 & 36.42 & 40.38 & 34.95 & 30.14 & 28.08 & 33.26 & 24.38 & \textbf{22.57} \\ 
GPT-4o & 49.22 & \textbf{75.76} & 51.93 & 23.12 & \textbf{45.56} & 51.62 & \textbf{43.84} & 32.00 & 35.73 & 44.61 & \textbf{30.76} & 19.93 \\
GPT-4o(PDF) & \textbf{55.70} & 66.67 & \textbf{58.27} & \textbf{36.02} & 45.14 & \textbf{53.24} & 41.30 & \textbf{38.00} & \textbf{37.08} & \textbf{49.39} & 29.30 & 19.71 \\
GPT4-turbo & 37.71 & 72.73 & 38.79 & 18.33 & 35.12 & 37.22 & 34.51 & 30.44 & 26.06 & 28.65 & 25.19 & 18.83 \\
Claude-3-Opus & 28.30 & 33.33 & 31.10 & 10.04 & 31.04 & 31.29 & 33.97 & 12.22 & 19.17 & 24.07 & 16.41 & 15.15 \\
Gemini-Pro & 25.79 & 48.48 & 26.91 & 11.29 & 17.20 & 19.19 & 16.29 & 15.07 & 10.22 & 12.74 & 8.90 & 5.03 \\
Baichuan-VL & 29.85 & 35.00 & 31.45 & 18.33 & 29.96 & 28.94 & 32.57 & 21.38 & 22.33 & 27.59 & 17.42 & 16.01 \\
Qwen-VL-Max & 34.82 & 42.86 & 36.65 & 20.45 & 24.87 & 25.70 & 24.96 & 20.72 & 16.95 & 18.97 & 15.61 & 14.92 \\
Qwen-VL-Plus & 20.49 & 40.00 & 21.23 & 9.20 & 19.16 & 21.11 & 18.83 & 13.19 & 11.00 & 13.94 & 9.29 & 5.83 \\
GLM4V & 10.32 & 33.29 & 9.62 & 4.29 & 13.28 & 17.07 & 14.85 & 12.89 & 7.64 & 8.73 & 11.11 & 4.08 \\
\midrule
\multicolumn{13}{c}{\textbf{Open-source Models (> 20B)}}  \\
\midrule
Qwen2-VL-72B & \textbf{46.72} & \textbf{56.06} & \textbf{49.46} & \textbf{26.34} & \textbf{42.15} & \textbf{45.33} & \textbf{42.33} & 26.42 & \textbf{30.77} & \textbf{45.43} & \textbf{19.34} & \textbf{19.04}\\
LLaVA-Onevision-72B & 7.27 & 6.06 & 8.08 & 2.69 & 9.16 & 12.20 & 7.07 & 11.05 & 7.77 & 7.17 & 8.71 & 6.09\\
LLaVA-Onevision-72B(PDF) & 36.98 & 72.73 & 37.60 & 20.0 & 34.10 & 39.11 & 31.90 & \textbf{27.46} & 23.41 & 30.85 & 17.57 & 17.37\\
Yi-VL-34B & 14.99 & 40.00 & 16.13 & 3.32 & 16.38 & 16.31 & 17.10 & 11.67 & 12.14 & 11.65 & 12.96 & 10.58 \\
\midrule
\multicolumn{13}{c}{\textbf{Open-source Models (< 20B)}}  \\
\midrule
DeepSeek-VL2-4.5B & 13.74 & 42.42 & 13.73 & 2.87 & 14.93 & 14.68 & 14.47 & 19.09 & 10.18 & 8.29 & 12.46 & 7.99 \\
LLaVA-1.6-7B & 9.77 & 35.21 & 10.82 & 7.12 & 10.37 & 9.79 & 10.90 & 9.07 & 7.57 & 8.41 & 6.53 & 4.54\\
Qwen2-VL-7B & \textbf{34.44} & \textbf{54.55} & \textbf{36.32} & 15.59 & \textbf{28.79} & \textbf{30.61} & \textbf{28.96} & 19.83 & \textbf{20.81} & \textbf{26.88} & \textbf{16.28} & 15.71\\
LLaVA-Onevision-7B & 13.80 & 9.09 & 14.49 & 11.29 & 19.56 & 20.11 & 19.01 & \textbf{21.20} & 14.59 & 14.28 & 15.86 & 9.86\\
Llama-3.2-11B-Vision & 24.68 & 30.30 & 25.56 & \textbf{17.20} & 20.78 & 20.79 & 21.86 & 13.00 & 19.15 & 23.33 & 16.13 & 15.06\\
Math-LLaVA-13B & 12.17 & 39.39 & 9.80 & 16.13 & 16.82 & 16.19 & 16.69 & 20.83 & 16.17 & 15.17 & 15.43 & \textbf{24.44}\\
G-LLaVA-7B & 8.56 & 6.06 & 9.30 & 4.84 & 15.96 & 17.74 & 14.82 & 16.50 & 13.03 & 15.69 & 11.42 & 8.97\\
\bottomrule
\end{tabular}%
}
\end{table*}

\begin{figure}
\centering
  \centering
  \includegraphics[width=0.49\textwidth]{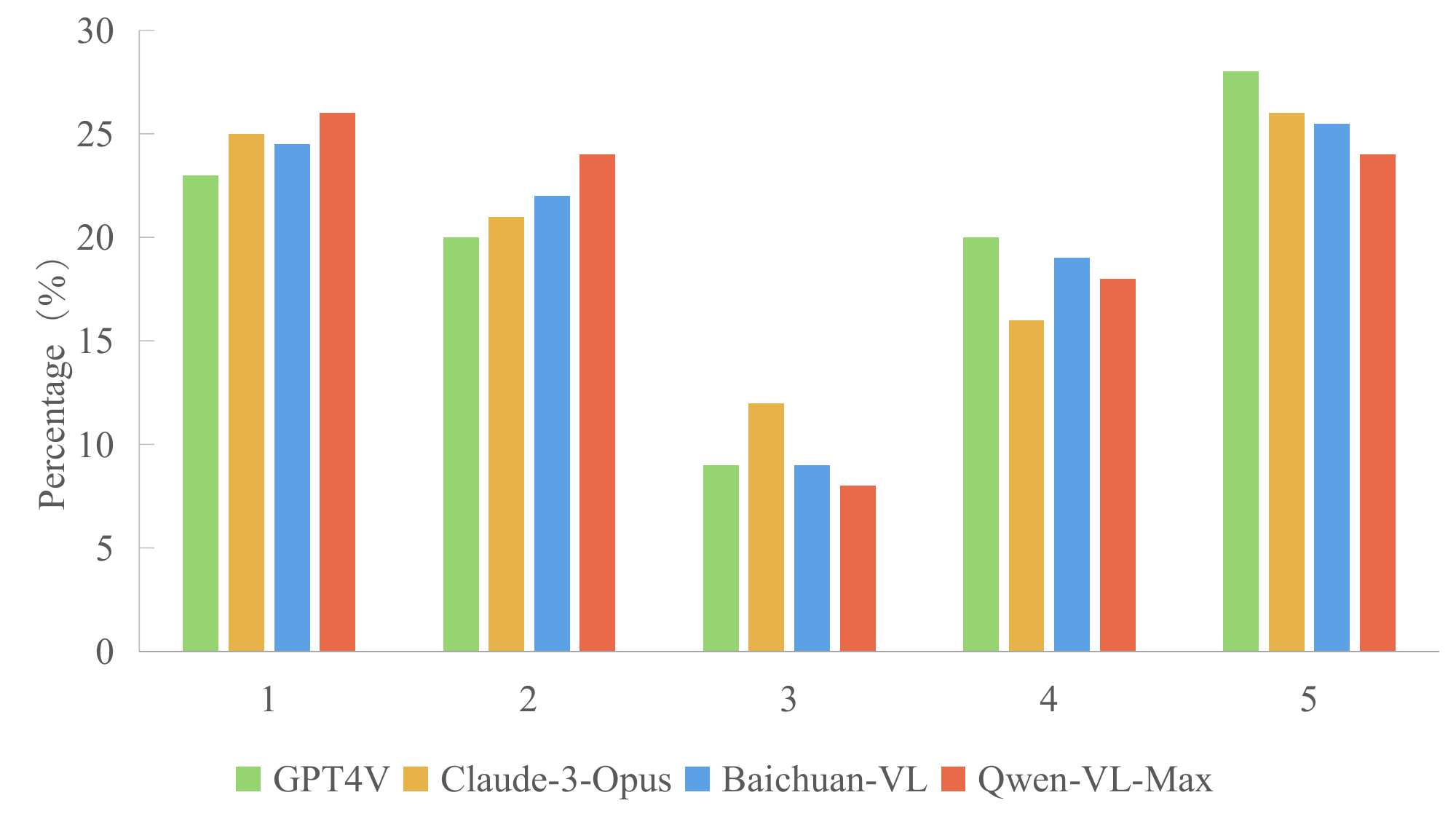}
  \caption{Each problem is tested five times, with the numbers 1 to 5 indicating the number of correct responses. Our results show that only about 25\% of the questions are answered correctly in all five attempts.}
  \label{fig: Stability}
\end{figure}

\subsection{The Importance of Real-World Images in Benchmarking}
To address \textbf{Q2}, we highlight the significance of real-world images in our MathScape benchmark. As shown in Tables \ref{tab:merged_1_2}, and \ref{Table 3}, \textbf{LLaVA-OneVision-72B} exhibits poor performance on the MathScape benchmark. However, when real-world images are replaced with clean PDF files, \textbf{LLaVA-OneVision-72B (PDF)} achieves a substantial performance improvement compared to \textbf{LLaVA-OneVision-72B}. This result demonstrates that high performance on clean PDF files does not necessarily translate to strong performance on real-world image settings. Furthermore, even SOTA models like \textbf{GPT-4o} experience a performance decline when evaluated on real-world images, as opposed to their performance on \textbf{GPT-4o (PDF)}. These findings emphasize the significance of real-world problems in our MathScape benchmark, which introduces novel challenges for MLLMs.

\subsection{Stability Results and Analysis}
To address \textbf{Q3}, we conducted a stability test on GPT-4V, Claude-3-Opus, Baichuan-VL, and Qwen-VL-Max. A total of 300 problems were selected, and each model was tested five times per problem. The number of correct answers across these attempts was recorded to evaluate stability. As shown in Figure~\ref{fig: Stability}, none of the models exhibited high stability—only around 25\% of the problems were answered correctly in all five attempts. This highlights the need to enhance the robustness of mathematical MLLMs to ensure more reliable performance in real-world applications.

\section{Conclusion}
In this work, we introduce MathScape, a new benchmark designed to assess the mathematical reasoning capabilities of MLLMs in real-world scenarios. Through comprehensive evaluations of both closed-source and open-source models, we highlight the challenges posed by real-world, photo-based mathematical problem-solving tasks. Our findings demonstrate that SOTA models, despite their strong performance on digitally rendered images, struggle significantly with real-world math problems. By introducing MathScape, we identify the strengths and weaknesses of current MLLMs, providing valuable insights into areas that require further improvement, particularly in reasoning, data robustness, and model generalization. We believe MathScape will serve as a critical benchmark to guide future research and the development of more capable MLLMs, enabling them to tackle complex, real-world mathematical challenges with greater accuracy and efficiency. Moving forward, we will explore more efficient and effective math multimodal data synthesize and annotation methods to enrich our benchmark.

\section{Acknowledgments}
This work is supported by the National Key R\&D Program of China (2024YFA1014003), National Natural Science Foundation of China (92470121, 62402016), and High-performance Computing Platform of Peking University.

\bibliographystyle{ACM-Reference-Format}
\balance 
\bibliography{main1}

\end{document}